\documentclass[10pt,twocolumn,letterpaper]{article}

\usepackage[pagenumbers]{cvpr} 

\usepackage{graphicx}
\usepackage{amsmath}
\usepackage{amssymb}
\usepackage{booktabs}
\usepackage{algorithm}
\usepackage{algorithmic}
\usepackage{comment}
\usepackage{multirow}
\usepackage{amsmath}
\usepackage{booktabs}
\usepackage{amssymb}
\usepackage[utf8]{inputenc}
\usepackage{color}
\usepackage{bm} 
\newcommand{\modelname}{{\usefont{T1}{ppl}{m}{n}CONE}}
\usepackage[pagebackref,breaklinks,colorlinks]{hyperref}

\usepackage[capitalize]{cleveref}
\crefname{section}{Sec.}{Secs.}
\Crefname{section}{Section}{Sections}
\Crefname{table}{Table}{Tables}
\crefname{table}{Tab.}{Tabs.}

\def\cvprPaperID{*****} 
\def\confName{CVPR}
\def\confYear{2023}

\begin{document}

\title{An Efficient COarse-to-fiNE Alignment Framework  @ Ego4D \\ Natural Language Queries
Challenge 2022}

\author{Zhijian Hou$^{1}$\thanks{\ \ \ Indicates equal contribution}, Wanjun Zhong$^{2*}$, Lei Ji$^3$, Difei Gao$^4$, Kun Yan$^5$ \\
Wing-Kwong Chan$^1$, Chong-Wah Ngo$^6$, Zheng Shou$^4$ and Nan Duan$^3$ \\
    $^1$ City University of Hong Kong \quad 
    $^2$ Sun Yat-sen University  \quad
    $^3$ Microsoft Research Asia \\
    $^4$ National University of Singapore \quad 
    $^5$ Beihang University \quad 
    $^6$ Singapore Management University  \\
    {\tt \small \{zjhou3-c@my., wkchan@\}cityu.edu.hk; 
     \tt \small zhongwj25@mail2.sysu.edu.cn; 
     \tt \small \{leiji, nanduan\}@microsoft.com} \\
    { \tt \small difei.gao@vipl.ict.ac.cn; 
      \tt \small kunyan@buaa.edu.cn;  
      \tt \small cwngo@smu.edu.sg;
      \tt \small mike.zheng.shou@gmail.com}\\
}

\maketitle

\begin{abstract}
This technical report describes the CONE~\cite{hou2022cone} approach for Ego4D Natural Language Queries (NLQ) Challenge in ECCV 2022. 
We leverage our model~\modelname, an efficient window-centric COarse-to-fiNE alignment framework. Specifically, \modelname~dynamically slices the long video into candidate windows via a sliding window approach.
Centering at windows, \modelname~(1) learns the inter-window (coarse-grained) semantic variance through contrastive learning and speeds up inference by pre-filtering the candidate windows relevant to the NL query, and (2) conducts intra-window (fine-grained) candidate moments ranking utilizing the powerful multi-modal alignment ability of the contrastive vision-text pre-trained model EgoVLP. On the blind test set, \modelname~achieves 15.26 and 9.24 for R1@IoU=0.3 and R1@IoU=0.5, respectively.

\end{abstract}

\section{Introduction}
Ego4D\cite{grauman2022ego4d} NLQ task aims to localize a temporal moment in a long first-person video according to a natural language (NL) question. 
It resembles the video temporal grounding task, since both share the same problem definition. However, statistics show the ratio of groundtruth to video duration is much smaller in Ego4d-NLQ compared with current video grounding benchmarks.
Specifically, Charades-STA and ActivityNet-Captions require the system to locate an 8s moment from a 30s video and a 36s moment from a 120s video on average, respectively. In contrast, the average groundtruth moment and video duration are 8 and 495 seconds in Ego4d-NLQ, making it a challenging “needle in the haystack” search problem.
Thus, directly adopting the existing methods for this task raises two major challenges:
(1) entire long videos with variant lengths are hard to be modelled without decreasing the sampling rate, leading to high computational cost during inference;
(2) a large number of moment candidates from long videos makes their precise multi-modal alignment with the NL query more challenging.  
These issues result in information loss in both \textit{temporal aspect} (i.e., fewer visible frames caused by down-sampling) and \textit{contextual aspect} (i.e., weaker matching ability to the object and scene in each frame content, as it is perturbed by numerous frames). 

To address the challenges above, we leverage our model~\modelname, a window-centric coarse-to-fine alignment framework. 
\cref{sec:method} illustrates the detailed method information. Result analysis~(\cref{sec:result_analysis}) reveals the effectiveness of CONE, and we end the report by discussing other challenges in first-person vision and temporal reasoning in the limitation analysis~(\cref{sec:limit_analysis}).

\begin{figure}[t]
\begin{center}
\includegraphics[width=0.45\textwidth]{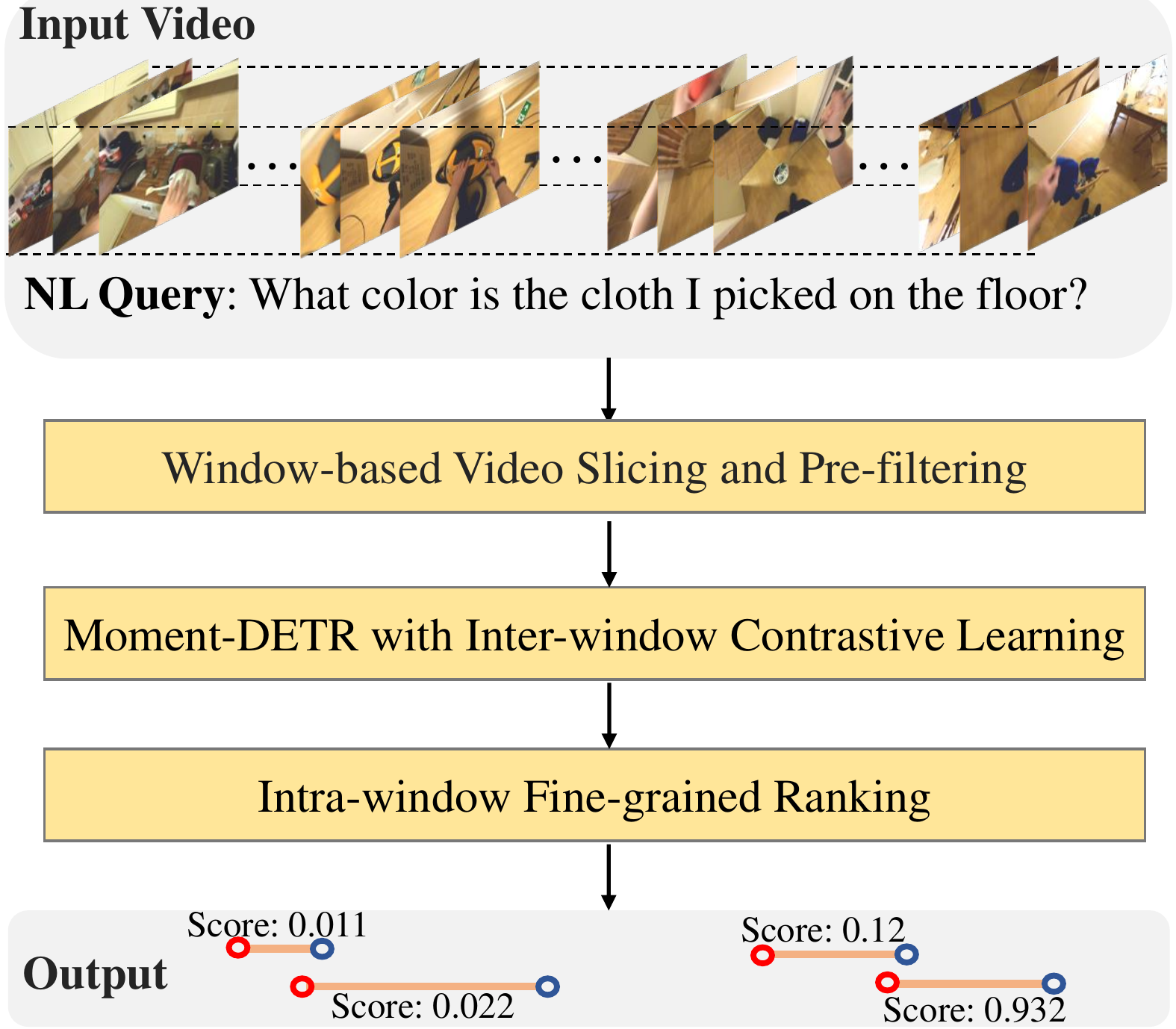}
\end{center}
\caption{An overview of our coarse-to-fine framework \modelname. \modelname~first slices the video with sliding window approach and accelerates inference with a pre-filtering mechanism (\cref{sec:window and pre-filtering}). At the coarse-grained level, it enhances training with inter-window contrastive learning (\cref{sec:contrast}). At the fine-grained level, it enhances fine-grained multi-modal alignment for accurate proposal ranking (\cref{sec:fine-grained}).}
\label{fig:cone}
\vspace{-0.2in}
\end{figure}

\section{Methodology}
\label{sec:method}
We present the proposed \modelname~for long video temporal grounding. 
As the pipeline shown in Fig.~\ref{fig:cone}, we first slice the long video into several fixed-length video windows using \textbf{sliding window} approach and \textbf{pre-filter} the candidate windows to accelerate inference (\cref{sec:window and pre-filtering}). 
Centering at the windows, we make accurate multi-modal alignment in the coarse-to-fine manner. At the coarse-grained level, we conduct \textbf{inter-window contrastive learning} (\cref{sec:contrast}) to capture inter-window semantic variance through training.
At the fine-grained level, we generate candidate intra-window proposals with \textbf{Moment-DETR}~\cite{lei2021detecting} and rank the proposals with a \textbf{fine-grained matching} (\cref{sec:fine-grained}) mechanism.

\subsection{Window-based Video Slicing and Pre-filtering}
\label{sec:window and pre-filtering}

We first slice the entire video into several video windows via a sliding window approach. A sliding window with window length $L_w$ is used to be slid on the entire video to derive a set of $N_w$ fixed-length video windows $W_i=[v_{w^b_i+1},v_{w^b_i+2},...,v_{w^b_i+L_{w}}]$, where $w^b_i$ is the start index of window $i$.
Specifically, we slide the window by increasing $w^b$ with window stride $L_{w}/2$ to guarantee that each moment is covered by two windows. 

Then, we propose to pre-filter the candidate windows by a contrastive vision-language pre-trained model EgoVLP~\cite{lin2022egocentric}.
We adopt it to compute the video features $\bm{V}$ and the text features $\bm{Q}$ beforehand. 
\begin{equation}
\label{equ:features}
 \begin{aligned}
  \bm{V}&=[\bm{v}_1,\bm{v}_2,\dots,\bm{v}_{L_v}] \\
  \bm{Q}&=[\bm{q}_{\texttt{[CLS]}},\bm{q}_1,\bm{q}_2,\dots,\bm{q}_{L_q}]
 \end{aligned}
\end{equation}
where \texttt{[CLS]} is a special token at the beginning of text tokens.

The multi-modal alignment score $a_{j} = \bm{v}_j \cdot \bm{q}_{\texttt{[CLS]}}$ is computed via the efficient dot product between $j^{th}$ video feature and the text feature. And the window-level matching score $A_{i}$ is the maximum score of all frames in window $i$: 
\begin{equation}
    A_{i} = \max([a_{w^b_i+1},a_{w^b_i+2},...,a_{w^b_i+L_w}])
\end{equation}

We rank all windows with $A_{i}$ and select the top-$k$ windows for inference. Thus, we reduce the number of candidate windows from $N_w$ to a constant $k$ to guarantee a controllable computation cost and accelerate inference.

\subsection{Inter-window Contrastive Learning}
\label{sec:contrast}
We conduct contrastive learning  to help the model in discriminating the semantic difference between positive and negative windows. 
Directly adopting Moment-DETR for long video in our framework is impractical because it lacks the ability to identify relevant windows, which hinders the reliable ranking of inter-window proposals. 

Contrastive learning is a feasible way to solve this issue.
In the scenario of temporal grounding, we expect the model to recognize the negative windows by giving lower confidence (saliency) scores to the proposals (frames) residing in the negative window:
\begin{equation}\label{eq:neg_loss}
\begin{aligned}
    p^+ &> p^- \\
    \textrm{mean}(S(W^{+},Q)) &> \max(S(W^{-},Q))
\end{aligned}
\end{equation}
where $p^{(+/-)}$ denotes the positive/negative proposal score. $W^{+}$ and $W^{-}$ denote the positive and negative window. S() is the saliency scoring function. 
Then, we design two-level contrastive losses: (1) proposal-level loss and (2) frame-level loss with a randomly sampled negative window. 

For the proposal-level contrastive loss ($\mathcal{L}_{p}$), we formulate it as a classification loss for predicting whether the window is relevant to the NL query:
\begin{equation}
    \mathcal{L}_{p} = - \sum \log \frac{\exp (p^+)}{\exp(p^+)+
    \exp(p^-)}
\end{equation}

For the frame-level loss $\mathcal{L}_f$, we set the average saliency scores for frames located in the positive window is larger than the maximum saliency score of frames in the negative window over a margin $\delta$:
\begin{equation}
\small
     \mathcal{L}_{f}=\max(0, \delta + \max(S(W^{-},Q))-\textrm{mean}(S(W^{+},Q))
\end{equation}
So the overall contrastive loss is $\mathcal{L}_c=L_{p}+L_{f}$. For the positive window, we also add the moment localization loss with L1 and IoU loss as in Moment-DETR.

\subsection{Intra-window Fine-grained Ranking}
\label{sec:fine-grained}
Moment-DETR exploits the self-attention mechanism to perform multi-modal fusion over the sequence of video and text features.
However, with the increased length of video inputs, the fine-grained attention between each video frame and the text query will be weakened by many other perturbed frames, resulting in \textit{contextual information loss}. And the fine-grained content (e.g., object and scene) in each frame plays an essential role.

To remedy this issue, we propose a novel ranking strategy to enhance the fine-grained multi-modal alignment with a matching score computed with a contrastive vision-text pre-trained model (described in \cref{sec:window and pre-filtering}).
Specifically, we first pre-compute the features for video frames and text query with the contrastive pre-trained model as in Eq. \eqref{equ:features}.

\paragraph{Visual Adapter.}
With a lightweight visual adapter on the top of EgoVLP, we adopt adapter-based tuning to adapt the representations from the general contrastive model to the data distribution of the current downstream task.

Our main idea is to add an additional bottleneck layer to learn the task-adaptive visual features and conduct residual-style blending with the original pre-trained
features. 
The lightweight adapter complies with a 2-layer FFN followed by ReLU. The $i^{th}$ adapted visual feature is: 
$\hat{\bm{v}_i} = \textrm{Adapter}(\bm{v}_i) + \bm{v}_i$.
Then, the \textbf{proposal feature} for the $j^{th}$ proposal is computed with the mean pooling of all the adapted video features in it: $\bm{h}_j=\textrm{Mean}([\hat{\bm{v}_{b_j}},\dots,\hat{\bm{v}_{e_j}}])$

For adapter training, we denote the positive proposal (with feature $\bm{h}_{pos}$) as the ground-truth one, and the negative proposals are the other in the same batch. We follow the standard contrastive learning and use the NCE loss.
\begin{equation}
    \mathcal{L}_{a} = - \sum_{pos} ( log \frac{\exp(\bm{h}_{pos}\cdot \bm{q}_{\texttt{[CLS]}})}{\sum_{j}\exp(\bm{h}_{j}\cdot \bm{q}_{\texttt{[CLS]}})} )
\end{equation}

\paragraph{Ranking Score Computation.}
Finally, we aim to conduct fine-grained ranking for proposals in the window. For the $j^{th}$ proposal, the final proposal ranking score is fused with two components: (1) proposal scores generated from Moment-DETR and (2) fine-grained matching score $m_j$  computed with CLIP-based proposal feature: $m_j=\bm{h}_j\cdot\bm{q}_{\texttt{[CLS]}}$. 
The former models the correlation between proposals by the Transformer-based architecture, while the latter focuses on fine-grained content matching between frames in the proposal and the text query. 
We perform min-max normalization for these two types of scores for a more stable ranking. 
The final ranking score $r_{j}$ is the sum of two normalized scores:
\begin{equation}
\label{eq:neg_loss}
\begin{aligned}
 \tilde{p_{j}} = &~\textit{MinMax}([p_1,p_2,...,p_{N_p}]) ,\\
 \tilde{m_{j}} = &~\textit{MinMax}([m_1,m_2,...,m_{N_p}]) \\ 
 r_{j} = &~\tilde{p_{j}} + \tilde{m_{j}} \\
\end{aligned} 
\end{equation}
where $N_p$ is the total number of candidate proposals.

\begin{figure*}[t]
\begin{center}
\includegraphics[width=\textwidth]{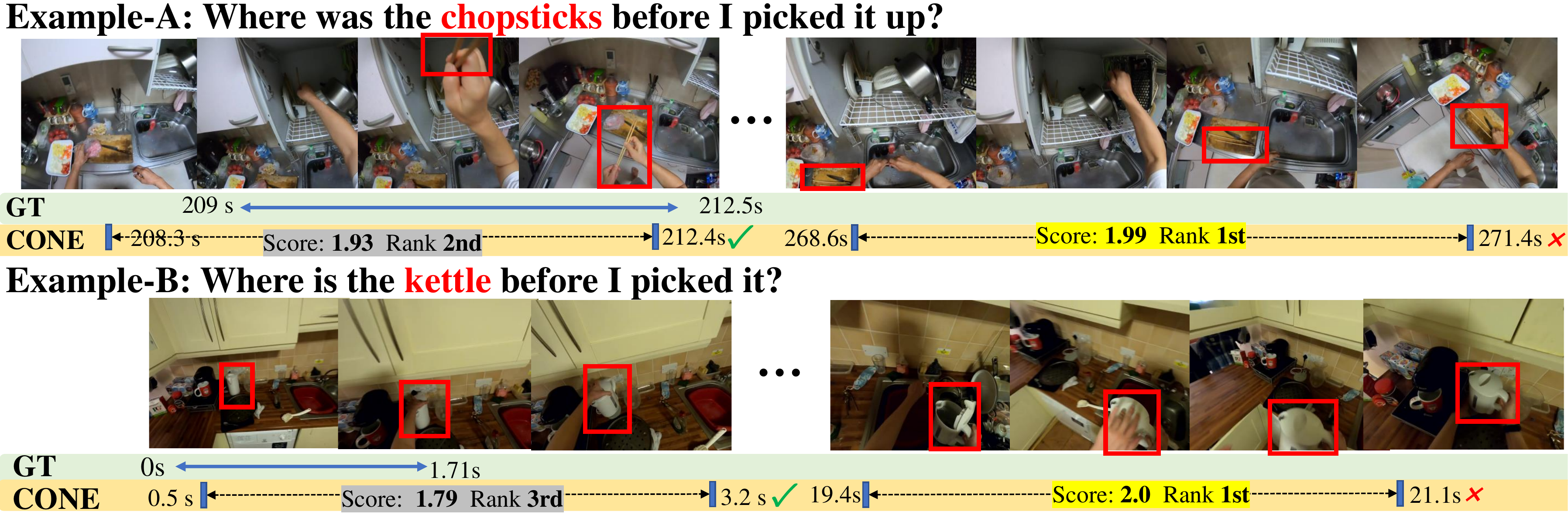}
\end{center}
\caption{Two failure cases. We show both groundtruth moment and our prediction moments.}
\label{fig:qualitative}
\vspace{-0.2in}
\end{figure*}

\section{Experiments}
\subsection{Implementation Details}

\paragraph{Visual, Textual and Token Features.}
We use EgoVLP\cite{lin2022egocentric} visual feature (every 16 frames, 1.87fps) and textual feature for all experiments.
They are utilized in the visual input of Moment-DETR, pre-filtering stage and fine-grained ranking stage. Moreover, we explore three different sources~(i.e., EgoVLP, RoBERTa and CLIP) to extract textual token feature.

\paragraph{Training and Inference Details.}
All experiments are conducted on one P100 GPU. Following Moment-DETR~\cite{lei2021detecting}, we set the hidden size $d$ to 256, the layer number in encoder/decoder to 2, and number of moment queries to 5 for the transformer module.
Parameter optimization is performed by AdamX. We set the learning rate to 1e-4 for Moment-DETR and 1e-5 for visual adapter, respectively. We set the batch size to 32 and adopt the early stopping strategy. 
We set the window length to 90 video features~(48 seconds) and train it for 150 epochs lasting about 3 hours.
We set the pre-filtering window number to 20 and use Non Maximum Suppression~(NMS) with a threshold of 0.5 as post-processing during inference.

\subsection{Result Analysis}
\label{sec:result_analysis}
\begin{table}[htbp]
  \centering
    \begin{tabular}{c|c|cc}
    \toprule
    \multicolumn{1}{c|}{\multirow{2}[2]{*}{Method}} & \multicolumn{1}{c|}{\multirow{2}[2]{*}{Textual Token Feature}} & \multicolumn{2}{c}{R1@IoU=} \\
          &       & 0.3   & 0.5 \\
    \midrule
    VSLNet & EgoVLP & 10.84  & 6.81 \\
    Moment-DETR & EgoVLP & 7.20  & 4.31 \\
    CONE  & EgoVLP & 14.15 & 8.18 \\
    CONE  & RoBERTa & 14.35 & 8.39 \\
    CONE  & CLIP  & 14.48 & 8.67 \\
    CONE  & Ensemble & 15.57 & 9.14 \\
    CONE (+SW) & Ensemble & \textbf{16.06} & \textbf{9.34} \\
    \bottomrule
    \end{tabular}%
    \caption{Performance~(Recall 1@IoU=$\theta$) on the validation set.}
  \label{table:ego4d}
\end{table}%

\begin{table}[htbp]
  \centering
    \begin{tabular}{l|cccc}
    \toprule
    \multicolumn{1}{c|}{\multirow{2}[2]{*}{Method}} & \multicolumn{2}{c}{{R1@IoU=}} & \multicolumn{2}{c}{{R5@IoU=}} \\
          & 0.3   & 0.5   & 0.3   & 0.5 \\
    \midrule
    CONE  & 15.26 & 9.24  & 26.42 & 16.51 \\
    \bottomrule
    \end{tabular}%
  \caption{Our final submission performance on the test set.}    
  \label{table:ego4d-test}%
\end{table}%

Table~\ref{table:ego4d} reports the performance comparison on the validation set. Under the same setting, CONE outperforms the performance of Moment-DETR~\cite{lei2021detecting} and VSLNet~\cite{zhang2020span} on both metrics by a noticeable margin.
It shows the effectiveness of the introduced inter-window contrastive learning mechanism and fine-grained ranking strategy.

Next, we study the effect on the token feature, which serves as the textual input of Moment-DETR. Better token features can result in more precise proposal boundary,
more discriminative proposal scores, and finally, better performance.
We empirically find that CLIP token feature works better than that of  RoBERTa and EgoVLP. We speculate that the reason lies in that CLIP is pre-trained on multi-modal corpus instead of RoBERTa text-only corpus, and the CLIP image-text pair corpus is much larger than EgoVLP first-person video-text pair corpus.

Then, we perform model ensemble on three CONE variants based on different token feature~(i.e., EgoVLP, RoBERTa and CLIP) and the ensemble strategy also shows a performance gain. 

Finally, we borrow the multi-scale variable-length sliding window sampling strategy~(SW)~\cite{liu2022reler} as a data augmentation trick. Previously, CONE slices the video into a sequence of pre-defined windows, and chooses a positive and negative window during training. In the additional SW version, the positive window has relatively random duration and start timestamp controlled by some hyperparameters.

For the final submission, we use our best version on the validation set. i.e., the ensemble of three CONE variants with the additional SW training strategy. 
As shown in Table~\ref{table:ego4d-test}, our submission achieves 15.26 and 9.24 for R1@IoU=0.3 and 0.5, respectively.


\subsection{Limitation Analysis}
\label{sec:limit_analysis}
Figure~\ref{fig:qualitative} shows two failure examples contrasting the groundtruth and predicted moments.
Both queries fail to rank in the first position but successfully rank at or before the fifth. We choose them because those queries are most likely to rank in the first position in future work.
In Example-A, CONE locates a moment when I am putting a peeler back. Although the chopsticks are visible on the chopping board, I am not interacting with them. In Example-B, CONE locates a moment when I am lifting the kettle from the sink, because the kettle has been filled with water from the tap. Although the kettle is the interaction object, I am not picking it up.
In contrast, the groundtruth shows a moment when I enter the kitchen and then pick up the kettle from the table.

CONE deals with the long video length challenge of the NLQ task. Nevertheless,
we speculate that other challenges also exist in first-person vision and temporal reasoning. Firstly, first-person vision requires the system to focus more on the interaction object~(shown by Example-A), and the most frequent query templates of the NLQ task~(i.e, ``Where is object X before / after event Y?'' and ``Where did I put X?'') all demand the necessity of human-object interaction. Secondly, temporal reasoning requires the system to understand more about the relation and causality of continuous actions or events~(shown by Example-B). This ability is key to this query template ``Where is object X before / after event Y?''.

{\small
\bibliographystyle{ieee_fullname}
\bibliography{egbib}
}

\end{document}